\documentclass[11pt,twoside]{article}
\usepackage{eacl2003}
\usepackage{times}
\usepackage{latexsym}
\usepackage{alltt}
\usepackage[english,ngerman,german]{babel}
\usepackage{graphicx}
\usepackage{array}
\title{Issues in Exploiting GermaNet as a Resource in Real Applications}

\author{Manuela Kunze and Dietmar R{\"o}sner\\
  Otto-von-Guericke-Universit\"at Magdeburg\\Institut f\"ur
Wissens- und Sprachverarbeitung \\
P.O.box 4120, 39016 Magdeburg\\
  {\tt makunze|roesner@iws.cs.uni-magdeburg.de}
}

\date{}

\begin{document}
\maketitle
\begin{abstract}
This paper reports about experiments with GermaNet as a resource
within domain specific document analysis. The main question to be
answered is: How is the coverage of GermaNet in a specific domain?
We report about results of a field test of GermaNet for analyses
of autopsy protocols and present a sketch about the integration of
GermaNet inside XDOC.\footnote{XDOC stands for \emph{X}ML based
\emph{doc}ument processing.} Our remarks will contribute to a
GermaNet user's wish list.
\end{abstract}

\section{Introduction}
GermaNet -- a lexical-semantic net -- was developed in the context
of the LSD-project: ``Ressourcen und Methoden zur
semantisch-lexikalischen Disambiguierung'' \cite{hinrichs:98}.
This paper describes an experiment about the integration of
GermaNet into the Document Suite XDOC. The Document Suite XDOC was
designed and implemented as a workbench for flexible processing of
electronically available documents in German
\cite{roesner.kunze:2002coling}.

We currently are experimenting with XDOC in a number of
application scenarios. These include:
\begin{itemize}
\item Knowledge acquisition from technical documentation about
casting technology as support for domain experts for the creation
of a domain specific knowledge base. \item Extraction of company
profiles from WWW pages for an effective search for products and
possible suppliers. \item Information extraction from English
Medline abstracts. \item Analysis of autopsy protocols for e.g.
statistical investigation of typical injuries in traffic
accidents.
\end{itemize}

The end users of our applications are domain experts (e.g. medical
doctors, engineers, ...). They are interested in getting their
problems solved but they are typically neither interested nor
trained in computational linguistics. Therefore the barrier to
overcome before they can use a computational linguistics or text
technology system should be as low as possible.

Many of our tools have an extensive need for linguistic resources.
Therefore we are interested in ways to exploit existing resources
with a minimum of extra work. The resources of GermaNet promise to
be helpful for different tasks in our workbench. GermaNet -- a
German version of the Princeton WordNet
\cite{miller:90,fellbaum:98} -- is based on the same design
principles, i.e. database structures like WordNet. The intention
of GermaNet is defined as the coverage of basic vocabulary of the
German language -- based on lemmatized frequency lists from text
corpora (see \cite{hamp.feldweg:1997} or \cite{kunze:2001}).

The following scenarios for the integration of the GermaNet
resource in our work are possible (see also section
\ref{sec-discussion}):

\begin{itemize}
    \item GermaNet as resource for semantic analyses,
    \item GermaNet for a shallow recognition of implicit document structures,
    \item GermaNet for compound analysis.
\end{itemize}

This paper is organized as follows: first we give a short
description of the document class `autopsy protocols', because the
examples of this paper are based on this corpus. After this we
describe our results related to the coverage of GermaNet for our
corpus and how the ambiguities inside the results can be resolved.
Then we shortly sketch the semantic module of XDOC into which the
resources of GermaNet should be integrated. This is followed by
the presentation and discussion of results from our experiments.
Our remarks will finally contribute to a GermaNet user's wish
list.

\subsection*{Characteristics of the Document Class: Autopsy
Protocols}\label{sec-autopsy} Autopsy protocols are especially
amenable to processing with techniques from computational
linguistics and knowledge representation:
\begin{itemize}
\item Forensic autopsy protocols are in most cases written with
the clear constraint that they will be used for legal purposes and
will have to be interpretable by lawyers and other non-medical
experts. \item Autopsy protocols are highly structured and follow
a strict ordering. \item The sub-language of the \emph{Findings}
section is of a telegramatic style with a  preference for
`verbless' structures. The sub-language of other subdocuments is
slightly more complex, but still limited due to the communicative
requirements (e.g. precision, \hyphenation{uni-que-ness}
uniqueness of expression, understandability for non-experts).
\end{itemize}

\section{GermaNet and Autopsy Protocols}
In the following we do report about on going experiments with a
corpus of currently approx. 600 autopsy protocols from Magdeburg.
The corpus will soon be extended with protocols from other
institutes for forensic medicine from all parts of Germany and
shall in the long run be representative for autopsy protocols from
all German speaking countries.

The central question for our experiments: Given a corpus with
texts from a uniform domain how is GermaNet's coverage as such,
i.e. without any investment in extending the available GermaNet
resources? We did not attempt lexical analysis of the tokens
derived from our test corpus, except comparison with exhaustive
lists of tokens from closed word classes of German function words,
connectors, prepositions, etc. This is reflecting the situation
when a corpus from a new domain is processed for the first time
and many domain terms are new and not covered in lexical
resources.
\subsection{Coverage of GermaNet}\label{subsec-coverage}
First experiments with GermaNet demonstrate the coverage of
GermaNet for autopsy protocols.
\begin{table}
  \centering \selectlanguage{english}
  \caption{Coverage for Different Document Parts.}\label{tab-coverage1}
  \small
  \doublerulesep 1pt
  \setlength{\extrarowheight}{2pt}
  \begin{tabular}{|l||c|c|c|}
  \hline
  document type & word types & match & percentage coverage  \\
  \hline\hline
  Findings & 17520 & 2591 & 14,78  \\\hline
  Background & 8124 & 2274 & 27,99  \\\hline
  Discussion & 8562 & 1862 & 21,74  \\
  \hline
\end{tabular}
  \normalsize
\end{table}
Table \ref{tab-coverage1} shows the coverage rates for the central
document parts of an autopsy protocol. For this evaluation we use
tokens, restricted by following parameters:
\begin{itemize}
    \item The candidates are not function words, like
    conjunctions, prepositions, etc. Only words that are potential
    candidates for nouns, adjectives and verbs are tested.
    \item In autopsy protocols some tokens are `implicit markup', e.g. enumerations of
    titles or paragraphs like `II.' in `II. Innere Besichtigung'.
    These tokens were excluded from the test.
    \item The length of a potential candidate was restricted to greater than
    three characters.
\end{itemize}
With these restrictions we reduce the number of different tokens
to be evaluated in section \emph{Findings} from 18492 to 17520, in
section  \emph{Background} from 8901 to 8124 and in section
\emph{Discussion} from 9198 to 8562 tokens.

The least coverage of GermaNet exists in the section
\emph{Findings}. This is not astonishing, because there we have
many domain specific terms (e.g., like `Thalamus',
`submandibularis', `Hirnkontusionen' or `Injektion'). In addition,
the medical doctors use their own (subjective) vocabulary for the
description of injuries or other findings, like `weichk{\"a}seartig',
`metallstecknadelkopfgro{\ss}e' or `teerstuhlartiger'. The best
coverage could be achieved in the section \emph{Background}. Here
we have many words from common language. This document part
describes the case history (e.g. details of a traffic accident).
We rarely find domain specific terms in this section. The section
\emph{Discussion}, which combines the results of the
\emph{Findings} section and the facts from the \emph{Background}
section, ranks in the middle with a 21,74 percentage.

A segmentation of the coverage rates into different word classes
is shown in table \ref{tab-coverage2}. In these data all hits are
counted, without distinction whether a GermaNet entry exists for
one or more word classes, therefore the sum of a row is greater
than the number of matches in table \ref{tab-coverage1}.

In the coverage summary the word class adverb is ignored, because
at the time of writing there are only two synsets for adverbs
available in the version of GermaNet and we got zero matches in
our corpus for these adverbs.
\begin{table}
  \centering \selectlanguage{english}
  \caption{Coverage for Different Word Classes.}\label{tab-coverage2}
  \small
  \doublerulesep 1pt
  \setlength{\extrarowheight}{2pt}
\begin{tabular}{|l||c|c|c|}
  \hline
  document type & nouns & verbs & adjectives \\
  \hline \hline
  Findings  & 1573 & 351 & 806 \\\hline
  Background  & 1622 & 328 & 465 \\\hline
  Discussion & 1162 & 322 & 483 \\
  \hline
\end{tabular}
  \normalsize
\end{table}

Related to the word class we have uniform results across
subdocuments, the largest coverage figure is for nouns, followed
by adjectives and verbs. The high ratio of adjectives in the
section \emph{Findings} is due to the high frequent usage of
adjectives in this section.

\subsection{Characteristic of Uncovered Terms}
The tokens that had no entry in GermaNet can be divided into two
classes. Beside the uncovered lexical terms (like `Rotor' or
`Klinge') we have a lot of specific terms, which could not be
covered by GermaNet. The analysis of these uncovered specific
terms, which negatively affect the results above, gives the
following classification.

\small
\begin{description}
    \item [measured values and ranges:] `2cm', `4-9', `120ml',
    \item [named entities:] `Beck', `Otto-von-Guericke-Universit"at', `Opel',
    `Salvator-Krankenhaus', `B269', `Zehringen-Sibbendorf',
    \item [truncations:] `-aussenseite', `-wischspuren',
    \item [compounds:] \hyphenation{Pla-stik-drei-punkt-sicher-heits-schlues-sel}`Plastikdreipunktsicherheitsschluessel',
    `Oberschenkelspiralmehrfragmentfraktur', `weisslich-gelblich-roetlich-fleckige',
    \item [inflected words:] `Armes', `besitzt', `entnommen',
    \item [misspellings:] `Herzmnuskulatur', `Herrrren-T-Shirt', `Todeseinritt'.
\end{description}
\normalsize The first category are non-lexical tokens. Depending
on the domain and text type their form and frequency is varying.
They cannot be expected to be covered by GermaNet and are best
treated with special recognizers (e.g. regular expressions).

All items of the first three categories can be preselected by
different preprocessing steps, like regular expressions or methods
for named entity recognition. The categories \emph{misspellings}
and \emph{inflected words} can only successful (in terms of
GermaNet) be preprocessed by a complex morphological component,
including recognition of inflected words and orthographic similar
words. For the processing of compounds in GermaNet it is possible
to use the resources of GermaNet itself (see section
\ref{sec-discussion}).

\section{Resolving Ambiguities}\label{subsec-ambiguities} In this
section we discuss approaches for resolving ambiguities. The
discussion is related to the kind of ambiguity. In our use of
GermaNet we found three types of ambiguities. Type one is an
ambiguity on the POS level -- whether the token to be analysed is
for example a noun or a verb. The second type occurs when more
than one sense exists for a word class. The last type is a
combination of the first two types.

\subsection{Part-of-Speech Ambiguity}\label{subsec-morpho}
Table \ref{tab-morph} shows the ratios of entries with
Part-of-Speech ambiguity.\footnote{in short: POS ambiguity}

The first row are results of counting all matches with more than
one word class per literal, the percentage rate related to all
matches is given in parentheses.

The rows 2 to 5 present the number of matches in which a specific
combination of word classes, e.g. noun and verb, occurs. The first
value in parentheses displays the percentage rate related to all
matches and the second value is the percentage rate related to all
matches with POS ambiguity.

In all three document parts the highest case of POS ambiguity
occurs between nouns and verbs. For example, the token `Herzens' in
the phrase `Gewicht des Herzens ...' will be interpreted in
GermaNet both as noun and as verb.

Due to the verbless style for this section it is not astonishing
that only in the section \emph{Findings} a similar high ratio is
given for the case `nouns and adjectives'.

It can be assumed, that a simple check of capitalisation of a
token can probably decrease the rates of POS ambiguity. Taking
sentence initial positions into account simple upper-/lowercase
distinction could decrease the rate of `noun-verb' or
`noun-adjective' matches.

\begin{table}
  \selectlanguage{english}
  \caption{POS Ambiguity.}\label{tab-morph}
  \scriptsize
  \doublerulesep 1pt
  \setlength{\extrarowheight}{2pt}
  \begin{tabular}{|m{1cm}||m{1.7cm}|m{1.6cm}|m{1.5cm}|}\hline

     & \small Findings & \small Background & \small Discussion\\
    \hline\hline
    different word classes & 139 (5,36) & 135 (5,93) & 104 (5,58) \\
    \hline
    N and V & 72 (2,77; 51,79)  & 89 (3,9; 65,9) & 71 (3,8; 68,2) \\\hline
    N and ADJ & 64 (2,47; 46,04) & 35 (1,15; 25,92) & 31 (1,66; 29,8) \\\hline
    V and ADJ & 3 (0,11; 2,15) & 5 (0,21; 3,7) & 1 (0,05; 0,96) \\\hline
    N, V and ADJ & 0 (0; 0 ) & 6 (0,26; 4,44) & 1 (0,05; 0,96) \\
    \hline
  \end{tabular}
\normalsize
\end{table}
Another approach is based on POS information about the tokens to
be analysed (using e.g. MORPHIX \cite{finkler.neumann:88}). With
this additional POS information we can directly decide which
information we want to retrieve in GermaNet. In addition, we can
also use a simple heuristic approach based on the information
about the document section. In the section \emph{Findings}
readings of adjectives can be preferred over readings as verbs.

\subsection{Sense Ambiguity}\label{subsec-sense}
\begin{table}
  \centering \selectlanguage{english}
  \caption{Sense Ambiguities.}\label{tab-sense}
  \small
  \doublerulesep 1pt
  \setlength{\extrarowheight}{2pt}
  \begin{tabular}{|l||r|r|}
    \hline
     & ratio & percentage\\
    \hline\hline
    Findings & 1034 & 39,95 \\\hline
    Background & 914  & 40,26 \\\hline
    Discussion & 823  & 44,27 \\
    \hline
  \end{tabular}
\normalsize
\end{table}
The average number of senses for a token of our corpus covered by
GermaNet is approx. 1,76.
The highest number we get is for verbs
with ca 3 senses (average numbers of senses for verbs: 3,18;
nouns: 1,49 and adjectives: 1,62). It is apparent that in many
cases GermaNet returns more than one sense for an entry. Table
\ref{tab-sense} shows the number of tokens with more than one
sense related to the different document parts.

A method for resolving the senses is the use of contextual
information. The specific structure of our documents (division in
three main parts) and  content related separation into these parts
allowed to exploit this information for the determination of the
most likely sense. As a start we use here the information of the
semantic fields of GermaNet. Experiments show (by majorities)
clear differences between the parts (see table \ref{tab-sf}).

Although the subdocuments may differ slightly in this respect
there is a strong preference for medical readings (senses) for
potentially ambiguous words in the corpus of autopsy protocols.
This is especially true for the subdocument with information about
the examination findings. The subdocument with the background is
the place where the expectation for medical senses seems to be
weakest.

Please note that words may have even conflicting `medical
readings'. `Blase' may be an organ (bladder) or an injury (e.g.,
caused by fire).

In the \emph{Findings} section the most frequent GermaNet
categories are `nomen.K\"orper', `verb.Ver\"anderung',
`verb.Lokation'. For resolving ambiguities we use this information
(majorities) for preselecting senses depending on the current
document section. For example, the noun `Becken' will be
classified by GermaNet in the semantic fields `nomen.Artefakt' (in
a sense of `music instrument') and `nomen.K\"orper' (in a sense of
`bone'). In the analysis of the \emph{Findings} section we prefer
the sense of `nomen.K\"orper'. In the section \emph{Background}
the sense of `nomen.Artefakt' has a higher likelihood than the
sense 'nomen.K\"orper'.

\begin{table}
  \centering \selectlanguage{english}
  \caption{Typical Semantic Fields of the Document Parts.}\label{tab-sf}
  \small
  \doublerulesep 1pt
  \setlength{\extrarowheight}{2pt}
  \begin{tabular}{|l||m{4cm}|}
    \hline
      section & most frequent semantic fields\\
    \hline\hline
    Findings & nomen.K\"orper, verb.Lokation, verb.Ver\"anderung, adj.K\"orper, adj.Perzeption,
    \\\hline
    Background & nomen.Geschehen, adj.Zeit, adj.Lokation \\\hline
    Discussion & nomen.Geschehen, nomen.K\"orper, verb.Lokation, verb.Ver\"anderung, adj.Relation \\
    \hline
  \end{tabular}
\normalsize
\end{table}

\subsection{Combined Ambiguity}
These cases are very rare in the corpus: \emph{Findings}: 11 (0,42
$\%$), \emph{Background}: 19 (0,83 $\%$) and \emph{Discussion}: 15
(0,8 $\%$). They could probably be resolved through the approaches
that are outlined in section \ref{subsec-morpho} and section
\ref{subsec-sense}.

\section{GermaNet inside the Semantic Module of XDOC}\label{sec-semantic}
The integration of the GermaNet resources takes place for the
purposes of semantic analysis. In this section we outline the
strategies for semantic analysis within XDOC. The \emph{Semantic
Module} in XDOC exploits three analysis techniques for the
annotation of documents with semantic information. The results of
the analysis are recorded in separate Topic Maps or annotated
within documents with a specific XML format. At first we give a
short description of the semantic analyses inside XDOC.

\paragraph{Semantic Tagger.}
The \emph{Semantic Tagger} classifies content words into their
semantic categories (different applications may have different
organizations of those categories in the form of taxonomies or
ontologies). For this function we expect as input data a text
tagged with POS tags and we then apply a semantic lexicon. This
lexicon contains the semantic interpretation of a token and a case
frame combined with the syntactic valence requirements.
\hyphenation{Si-mi-lar}Similar to POS tagging, the tokens in the
input are annotated with their meanings and with a classification
into semantic categories (i.e. specific concepts or relations). It
is possible that the classification of a token in isolation is not
unique. In analogy to the POS tagger, a semantic tagger that
processes isolated tokens is not able to disambiguate between
multiple semantic
\hyphenation{ca-te-go-ri-sa-tions}categorisations. This task is
postponed for contextual processing within case frame analysis
(\emph{Semantic Parser}).

\begin{figure}[hbtp]
\begin{center}
\includegraphics[scale=0.4]{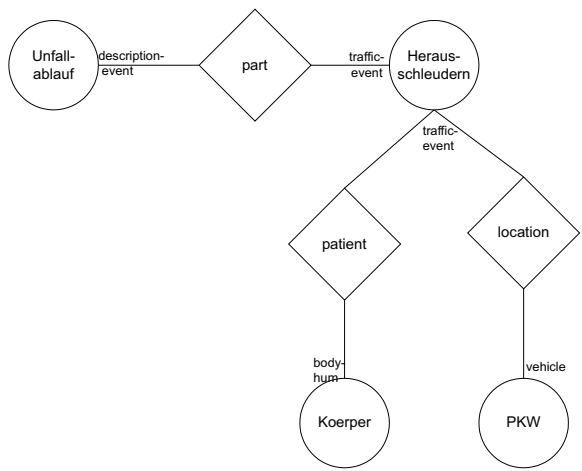}
\selectlanguage{english} \caption{Results of Case Frame
Analysis.}\label{case}
\end{center}
\end{figure}

\paragraph{Semantic Parser.}
The \emph{Semantic Parser} is one method in XDOC for the
assignment of semantic relations between isolated (but related)
tokens. By case frame analysis of a token we obtain details about
the type of recognized concepts (resolving multiple
interpretations) and possible relations to other concepts. Fig.
\ref{case} contains the results of the analysis of the noun phrase
\hyphenation{Un-fall-ab-lauf}\emph{`Unfallablauf mit
Herausschleudern der Koerper aus dem PKW'}. We get here the
assignments of the relation \emph{part} between
\emph{`Unfallablauf'} and \emph{`Herausschleudern der Koerper aus
dem PKW'} and the relations \emph{location} (between
\emph{`Herausschleudern'} and \emph{`PKW'}) and \emph{patient}
(between \emph{`Herausschleudern'} and \emph{`Koerper'}).

\begin{figure}[hbtp]
\begin{center}
\includegraphics[scale=0.4]{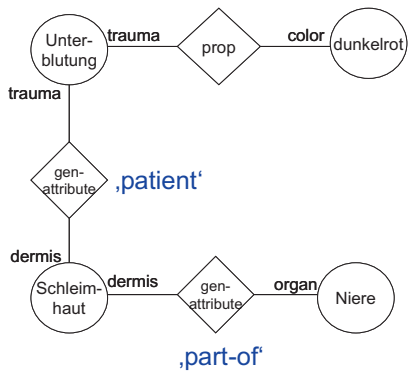}
\selectlanguage{english} \caption{Results of SIsS.}\label{siss}
\end{center}
\end{figure}

\paragraph{Semantic Interpretation of Syntactic Structure (SIsS).}
An other step for the recognition of relations between tokens is
the \emph{Semantic Interpretation of syntactic Structure} of a
phrase or sentence respectively. We exploit the syntactic
structure of the language (e.g., structures of noun phrases) and
the semantic interpretation of tokens inside the structure to
extract relations between several tokens. Fig. \ref{siss} is a
visualization of the results of the analysis of the noun phrase
\emph{`dunkelrote Unterblutung der Schleimhaut der Niere'}. The
analysis of this complex noun phrase results in three relations
between the separated nouns. The relation \emph{prop} is used to
label properties of a concept. Our future work here: The generic
relation \emph{gen-attribute} (short for attribute based on a
genitive surface case) has to be resolved into the appropriate
more specific relations, like \emph{part-of} or \emph{patient}.

The core of all these semantic analyses techniques is a semantic
lexicon. This lexicon records the meanings and case frames (only
for nouns and verbs) of a word.

Up to now the entries of this lexicon have been manually built up
and are partially domain dependent. Now we want to integrate the
GermaNet resources into our framework.

\subsection{Integration of GermaNet}\label{sec-germanetinsemantic}
Currently the integration of GermaNet is realised in the semantic
tagger. For the semantic lexicon we use the conceptual relation
\emph{hypernym} of GermaNet. The tagger uses the first level of
the \emph{hypernym} relation for the annotation of tokens with
information about the GermaNet senses:

\scriptsize
\begin{verbatim}
(tag-semantic-xml "<N>Leber</N> <S-KONJ>und</S-KONJ>
<N>Niere</N>")

"<CONCEPT TYPE="Innerei; Verdauungsorgan">Leber
</CONCEPT> <XXX><S-KONJ>und</S-KONJ></XXX>
<CONCEPT TYPE="Innerei; Harnorgan">Niere </CONCEPT>"
\end{verbatim}
\normalsize

The XML-attribute \emph{TYPE} contains the \emph{hypernym}
information from GermaNet. The different senses are separated by a
semicolon.

For better results we reconfigured our semantic tagger. In
contrast to the early version the semantic tagger now also expects
tokens with POS information (word classes), but enriched with
additional information about the stem of the tokens. In this way
we ask for senses related to a word class with the facility to use
a non-inflected word form for the request.

\scriptsize
\begin{verbatim}
(tag-semantic-xml "<N STEM="Gewicht">Gewicht</N>
<DETD>des</DETD><N STEM="Herz">Herzens</N>")

"<CONCEPT TYPE="?physikalisches Attribut; Wichtigkeit
Messgeraet, Messgeraet*o, Messinstrument*o,
Messinstrument; Artefakt, Werk">Gewicht</CONCEPT>
<XXX><DETD>des</DETD></XXX> <CONCEPT TYPE="Innerei;
Organ; Farbe, Spielfarbe; Flaeche, Ebene">Herzens
</CONCEPT>"
\end{verbatim}
\normalsize

Another integration of the GermaNet resources is possible for the
\emph{Semantic Parser}. Here we could use the information of verb
frames. Up to now the mapping of GermaNet verb frames to the XDOC
case frames could be problematic. For  case frames we use in
addition to syntactical valency (e.g., noun phrase in accusative)
also the description of potential semantic roles for the filler of
the frame. This information is not available from GermaNet's verb
frames. For this integration of GermaNet it is necessary to
complete the additional semantic information manually or by a
corpus based approach (learning from corpora). For instance, for
the analysis of the sentence `Sie wurde am Kopf operiert.' we get
for the verb `operieren' the GermaNet sense:

\scriptsize
\begin{alltt}
Sense 1 operieren
       => medizinisch behandeln
          => wandeln, {\"a}ndern, mutieren, ver{\"a}ndern
\end{alltt}
\normalsize
GermaNet contains for this sense following verb
frames:

\scriptsize
\begin{alltt}
Sense 1
operieren
          *> NN.AN
          *> NN.AN.BL
\end{alltt}
\normalsize

The second verb frame matches our example sentence.

But the usage of these GermaNet's verb frames in the analysis of
the sentence `Sie wurde im KKH xxx am Arm operiert.' is
problematic because the \emph{BL} complement could be assigned to
the locative preposition phrase `im KKH xxx' or to the locative
preposition phrase 'am Arm'. One of the two prepositional
complements gets no direct assignment to a complement defined by
GermaNet's verb frames. Other similar problematic examples from
our corpus are:

\small
\begin{itemize}
    \item \emph{Nach polizeilichen Angaben aus der Akte und den klinischen Unterlagen wurde G xxx/xx am
    Morgen des dd.mm.jj im Krankenhaus X wegen einer knotigen Kropfbildung operiert (Strumaresektion).}
    \item \emph{Am dd.mm.jj wurde G xxx/xx im KKH xxx am Herzen operiert.}
\end{itemize}
\normalsize
 A detailed description, e.g. additional information
about the semantic role of the complement's content, could be
helpful for the analysis. Our \emph{Semantic Parser} works with
such information. For the usage of the verb frames for the
analysis with our \emph{Semantic Parser} we need additional
features for the \emph{Adverbial Complement (BL)} of the verb
frame:\footnote{When we assumed that \emph{BL} is a preposition
phrase.}
\begin{itemize}
    \item semantic role of the filler: body part, for example, organs or extremities,
    \item possible preposition: am,
    \item case of PP: dative or not specified.\footnote{When no unique assignment to one case is possible.}
\end{itemize}

Other features to be considered in using verb frames are:
\begin{itemize}
    \item the different complement forms for active or passive usage of a
    verb and
    \item the number of a noun phrase: For example, for the verb
    `kollidieren' is the possible verb frame `NN.Pp'. The
    preposition phrase is defined as an optional complement.
    A necessary additional feature for the noun phrase is the
    information about its number (singular or plural).
    For example, the subject noun phrase in sentences like `Die Fahrzeuge
    kollidierten.' must name more than one participant of the accident.
\end{itemize}

To complete GermaNet's verb frames it is possible on the one hand
to add this additional information manually or on the other hand
by the analysis of occurrences of similar phrases in the corpus.
By the corpus based approach the user gets a list of possible
complements for a verb, so that the verb frame of GermaNet can be
enriched with the corpus/context related features. GermaNet's verb
frames are used as pattern for the search inside the corpus. The
basis for this approach is a corpus with syntactic structures
annotated by the \emph{Syntactic Parser} of XDOC
\cite{konvens2000}.

\begin{figure}[hbtp]
\begin{center}
\includegraphics[scale=0.35]{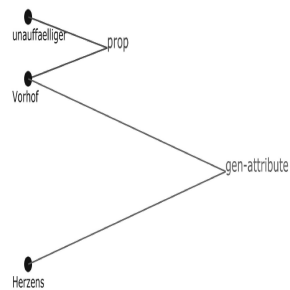}
\selectlanguage{english} \caption{Result for the Phrase
`unauffaelliger Vorhof des Herzens'.}\label{siss2}
\end{center}
\end{figure}

One problem inside the SIsS analysis is the correct interpretation
of the genitive-relation. One solution is the usage of the
conceptual relations \emph{meronym} and \emph{holonym} of
GermaNet. For example, the results of  the SIsS analysis of the
phrase 'unauffaelliger Vorhof des Herzens' is shown in Fig.
\ref{siss2}.

By the SIsS analysis two relations were recognised: first the
\emph{prop} relation between the tokens `unauffaellig' and
`Vorhof', the second recognised relation is the
\emph{gen-attribute} relation.  This relation can in general be
interpreted in several ways (e.g. as \emph{part-of} or
\emph{patient-of}). GermaNet results for the token `Herz' (in the
sense of \emph{organ}) give the meronyms: \emph{Vorhof,
Herzklappe, Herzkammer, Herzrohr, linke Herzh"alfte, rechte
Herzh"alfte, Herzmuskel, Herzkranzgef"a\ss} and for the token
`Vorhof' the holonym `Herz'. With \emph{meron} and \emph{holon}
information from GermaNet we can decide that the generic relation
(`gen-attribute') between the tokens `Herz' and `Vorhof' is a
\emph{part-of} relation.

\subsection{Practical Aspects of the Integration of GermaNet}
The technical access to GermaNet was realised in different ways:
offline and online usage of GermaNet. In offline usage GermaNet is
transformed into an application specific resource. This
transformation may be carried out as a compilation step
beforehand. Online usage employs GermaNet resources via their API.

For the offline usage of GermaNet we only transform necessary
information into the application specific resources. Depending on
the task to be performed we do need different information from
GermaNet. In one case we need the synsets and hypernyms
(\emph{Semantic Tagger)}, in other cases we only work with
information about the semantic fields of a token (for example,
\emph{scenario: shallow recognition of document sections}). The
relations inside GermaNet can also be described as path direction
-- up, down, horizontal (see \cite{hirst.st-onge:1998}). Within
the semantic module of XDOC the following `path directions' could
be useful:

\small
\begin{description}
    \item [Semantic Tagging:] search for a context-based allowed sense
    (hypernyms, synsets),
    \item [Semantic Parser:] assignment to semantic roles, search for a filler of a semantic
role (hypernyms), syntactic information via verb frames,
    \item [SIsS:] resolving of semantic interpretation of genitive-structures
    (e.g. Schleimhaut des Magens) by `meron' or `holon' information.
\end{description}
\normalsize
 Concerning the coverage of GermaNet we obtained the
following results:
\begin{itemize}
    \item Some tokens of our corpus are not covered by GermaNet,
    especially in the range of open word classes, like adjectives (e.g.
    `quer') or in the range of domain specific words (e.g.
    `Fraktur').
    \item Uncomplete senses of an entry. For instance, for the word
    `Abfall' there exists only one sense related to the semantic field
    `noun.substance', but in our corpus we often find the word
    `Abfall' in phrases like `Abfall des Blutdruckes'. Please
    note: For the verb `abfallen',
    from which the noun `Abfall' is derivated by
    a verb-noun-conversion, we also do not get the
    right sense related to our domain:

    \scriptsize
       \begin{alltt}
       1 sense of abfallen
       Sense 1 abfallen
       => l{\"o}sen
           => ?Dauerkontakt
      \end{alltt}
     \normalsize
     \end{itemize}
    From a more technical perspective the following points are relevant:
    \begin{itemize}
    \item A very simple or no morphological component in GermaNet (WordNet is better),
    e.g. `Autos' will be found, but `Organe' will not be found in
    GermaNet. This is explainable only through the use of an English
    morphology component (from WordNet). GermaNet uses English
    flection criteria for the analysis of the input data. By reconfiguration of our
    \emph{Semantic Tagger} we can avoid this effect (see section
    \ref{sec-germanetinsemantic}).
    \item Use of umlauts in GermaNet: The documents in our corpus are without umlauts, but
    GermaNet supports only access via writings with umlauts. Matching of
    candidates without umlauts to possible candidates in GermaNet
    with umlauts could be helpful and would lead to a better
    coverage.
\end{itemize}

In consideration of the last two points we worked with two
additional intermediate steps in our experiment environment. At
first we integrated the morphological component MORPHIX (revised
results for the different sections are \emph{Background:} 41,39
\%, \emph{Findings:} 29,64 \%, \emph{Discussion:} 40,57 \%) and
the second step was the treatment of umlauts which again improved
our GermaNet coverage results (\emph{Background:} 43,38 \%,
\emph{Findings:} 31,02 \%, \emph{Discussion:} 42,38 \%).

\subsection{User's Wish List}
Some items for the GermaNet user's wish list:
\begin{itemize}
    \item It seems that in the case of orthographic variants
    GermaNet `knows' sometimes more than it makes available. An
    example: GermaNet has the information that `4-eckig' is an
    orthographic variant of `viereckig', but does only return
    information when the user (or application program) asks
    with the (canonic) writing 'viereckig'.
    \item Flexible match of umlauts and extended writings: Given
    the fact that in computer written text umlauts are still often
    represented in the expanded form of `ae', `oe', \dots it would
    be helpful to increase the flexibility of GermaNet's lexicon
    access and provide means that search terms in the expanded
    writing will match  existing entries with umlauts (i.e.
    `Gebaeude' should match `Geb"aude').
    \item Avoid artefacts due to English spelling rules from
    Wordnet: Wordnet and GermaNet offer convenience
    functions to the user for search in the resources in the sense
    that some but not all inflections, derivations, and alternative spellings
    can be handled. For example: `Herzens' matches the verb
    `herzen'(!) but not the noun `Herz'.\footnote{Please note: `Herzens' can
    be erroneously derived from the verb `herzen' under the assumption of an English
    inflection: `English' morphological attributes of `Herzens' are then third person
    singular.}
    \item Finally: GermaNet is not error free. In our work we occasionally
get messages like `Error Cycle detected' or `Synset xxxx not
found', which make the user insecure about the results returned by
GermaNet.
\end{itemize}

\section{Discussion: Back to the Scenarios}\label{sec-discussion}
In previous sections we have described some integration aspects of
GermaNet for different scenarios. Now we give a concrete outline
of the scenarios.

\paragraph {GermaNet as Resource for Semantic Analyses.}
In section \ref{sec-germanetinsemantic} we described the
integration of GermaNet as resource in the \emph{Semantic Module}
of XDOC. There we use the lexical-semantic net for the annotation
of tokens with their semantic roles (\emph{Semantic Tagger}). For
this task we exploit the different defined relations inside
GermaNet (e.g. hypernym or synonym). For the tasks of the
\emph{Semantic Parser} and the \emph{SIsS analysis} we
additionally use information of verb frames and other conceptual
relations, like the `meron' and the `holon' relation. The
\emph{Semantic Parser} directly uses this information for the
analysis, while the \emph{SIsS analysis} uses GermaNet's
information in a postprocessing step for the selection of one
(possible) interpretation of the different readings resulting from
the \emph{SIsS analysis} (e.g. the relation \emph{gen-attribute}).

\paragraph {GermaNet for a Shallow Recognition of Implicit Document Structures.}
In section \ref{sec-autopsy} we have given a short specification
of autopsy protocols. The characteristics of the different
document parts can be used for a recognition of these parts. The
following parameters describe the different document parts (also
related to the available information by GermaNet):

\small
\begin{description}
    \item[Findings:] high ratio of nouns and adjectives; short specific syntactic (sentence)
    structures; semantic fields like `nomen.K{\"o}rper', `adj.K{\"o}rper',
    `verb.Ver{\"a}nderung',
    \item[Background] standard distribution of all word classes;
    regular syntactic structures; semantic fields like
    `nomen.Geschehen', `adj.Zeit', `adj.Lokation',
    \item[Discussion] standard distribution of all word classes;
    regular syntactic structures; semantic fields like
    `nomen.Geschehen', `nomen.K{\"o}rper', `verb.Lokation',
    `verb.Ver{\"a}nderung'.
\end{description}
\normalsize The distribution of the semantic fields over different
document parts can be used for the recognition of these document
parts. For example a document part with a high frequent occurrence
of tokens, which can be assigned to the semantic fields like
'nomen.Geschehen', 'adj.Zeit', 'adj.Lokation', and no occurrences
of tokens with assignments of 'nomen.K{\"o}rper' etc. can be
identified as the \emph{Background} section of an autopsy
protocol. For a unique identification we also use information
about the word classes by the \emph{POS Tagger} and the
information about the kind of syntactic structures by the
\emph{Syntactic Parser} to confirm the other characteristic
criteria of a document part.

\paragraph {GermaNet for Compound Analysis.} \label{subsec-compounds}
    In the autopsy protocol corpus -- as well as in other medical or technical
texts -- noun compounds are quite frequent. The question here is:
Is it possible to
\begin{itemize}
    \item safely determine segmentations of noun compounds and to
    \item construct
meaning hypotheses for noun compounds by combining the meaning of
the compound's parts if they are covered by GermaNet?
\end{itemize}

Please note: Segmentation of German noun compounds (i.e.
determination of boundaries between parts of a noun or noun
compound) may produce artefacts even when the hypothesized
compound segments are lexical entries in their own right.

Examples (suggested segmentations indicated with [ ... ]):

\scriptsize
\begin{alltt}
  Transport ... * [Tran][sport]
  Lebertransport ... * [Lebertran][sport]
                       [Leber][transport]
\end{alltt}
\normalsize

We therefore favour an approach to compound segmentation that
additionally takes the corpus and the occurrence frequencies of
complex words with common pre- and suffixes into account and thus
reduces the dependence on the lexicon and its coverage.

The corpus-based analysis of compounds with GermaNet can be
described as follows: The first step is to find all compounds with
similar suffixes inside the corpus, like `Nierentransplantation',
`Lebertransplantation' etc. Then define `Top Level' relations
between possible candidates for compounds, for our example:
$<$organs$>$$<$medical-operation$>$, to avoid a wrong
interpretation of compounds. Here we can use the semantic field
information of GermaNet for the description of relations between
possible candidates.

\section{Conclusion}
We have reported about first experiments in integrating GermaNet
resources into XDOC for the processing of autopsy protocols.

Although our results related to the coverage of GermaNet were not
as high as in Saito's experiments
\cite{saito.wagner.katz.reuter.burke.reinhard:2002}, the results
for a corpus of autopsy protocols are encouraging. (A parallel
experiment with the EUROPARL corpus -- available at
http://www.isi.edu/$\sim$koehn -- resulted in a lower coverage. Of
198546 tested tokens only 30344 tokens are covered by GermaNet;
this probably is in part due to the high ratio of named entities
in the Europarl corpus.) The results could be further improved by
XDOCs preprocessing steps, like named entity recognition, POS
tagger etc., so that an adoption of GermaNet resources into the
semantic analyses of XDOC is conceivable.

We use GermaNet's lexical-semantic net for semantic enrichment of
documents. GermaNet's resources were primarily integrated into the
\emph{Semantic Tagger} of XDOC. In future work we will further
extend the integration of GermaNet for the \emph{SIsS} analysis
and the \emph{Semantic Parser}.

\bibliographystyle{acl}
\bibliography{eacl2003}

\begin{thebibliography}{}

\bibitem[\protect\citename{Hinrichs et.al.}1998]{hinrichs:98}
E.~Hinrichs et.al.
\newblock 1998.
\newblock {LSD-Projekt im Forschungsschwerpunkt: Methoden und Ressourcen der
  lexikalisch-semantischen Disambiguierung}.
\newblock Abschlu{\ss}bericht.

\bibitem[\protect\citename{Fellbaum}1998]{fellbaum:98}
C.~Fellbaum.
\newblock 1998.
\newblock {\em {W}ord{N}et: {A}n {E}lectronic {L}exical {D}atabase}.
\newblock Mass.: MIT Press.

\bibitem[\protect\citename{Finkler and Neumann}1988]{finkler.neumann:88}
W.~Finkler and G.~Neumann.
\newblock 1988.
\newblock {MORPHIX:} a fast {R}ealization of a {classification-based}
  {Approach} to {Morphology}.
\newblock In H.~Trost, editor, {\em Proc. der 4. {\"O}sterreichischen
  {Artificial-Intelligence} {T}agung, {W}iener {W}orkshop {W}issensbasierte
  {S}prachverarbeitung}, pages 11--19. Springer Verlag, August.

\bibitem[\protect\citename{Hamp and Feldweg}1997]{hamp.feldweg:1997}
B.~Hamp and H.~Feldweg.
\newblock 1997.
\newblock {G}erma{N}et -- a lexical-semantic {N}et for {G}erman.
\newblock In P.~Vossen et.al., editor, {\em {Proc. of ACL/EACL-97 workshop
  Automatic Information Extraction and Building of Lexical Semantic Resources
  for NLP Applications}}, pages 9--15, Madrid.

\bibitem[\protect\citename{Hirst and St-Onge}1998]{hirst.st-onge:1998}
G.~Hirst and D.~St-Onge, 1998.
\newblock {\em {W}ord{N}et: {A}n {E}lectronic {L}exical {D}atabase}, chapter
  14. {L}exical {C}hains as {R}epresentations of {C}ontext for the {D}etection
  and {C}orrection of {M}alapropisms, pages 305--333.
\newblock Mass.: MIT Press.

\bibitem[\protect\citename{Kunze}2001]{kunze:2001}
C.~Kunze, 2001.
\newblock {\em {L}exikalisch-semantische {W}ortnetze}, pages 386--393.
\newblock Spektrum, Akademischer Verlag, Heidelberg; Berlin.

\bibitem[\protect\citename{Miller}1990]{miller:90}
G.~Miller.
\newblock 1990.
\newblock {F}ive {P}apers on {W}ord{N}et.
\newblock In {\em CSL-Report}, volume~43, Princeton University. Cognitive
  Science Laboratory.

\bibitem[\protect\citename{R{\"o}sner and Kunze}2002]{roesner.kunze:2002coling}
D.~R{\"o}sner and M.~Kunze.
\newblock 2002.
\newblock {A}n {XML} based {D}ocument {S}uite.
\newblock In {\em Coling 2002}, pages 1278--1282, Taipei, Taiwan, August.

\bibitem[\protect\citename{R{\"o}sner}2000]{konvens2000}
D.~R{\"o}sner.
\newblock 2000.
\newblock {C}ombining {R}obust {P}arsing and {L}exical {A}cquisition in the
  {XDOC} {S}ystem.
\newblock In {\em KONVENS 2000 Sprachkommunikation}, ITG-Fachbericht 161, ISBN
  3-8007-2564-9, pages 75--80. VDE Verlag, Berlin, Offenbach.

\bibitem[\protect\citename{Saito \bgroup et al.\egroup
  }2002]{saito.wagner.katz.reuter.burke.reinhard:2002}
J.-T. Saito, J.~Wagner, G.~Katz, P.~Reuter, M.~Burke, and S.~Reinhard.
\newblock 2002.
\newblock {E}valuation of {G}erma{N}et: {P}roblem {U}sing {G}erma{N}et for
  {A}utomatic {W}ord {S}ense {D}isambiguation.
\newblock In {\em {Proc. of the LREC Workshop on WordNet Structure and
  Standardization and how these Affect WordNet Applications and Evaluation}},
  pages 14--29, Las Palmas de Gran Canaria.

\end{thebibliography}

\end{document}